# Do Deep Learning Models and News Headlines Outperform Conventional Prediction Techniques on Forex Data?

Sucharita Atha[1], Bharath Kumar Bolla[2]

[1]Liverpool John Moores University, Rodney House, Liverpool L3 5UX, United Kingdom
[1]UpGrad Education Private Limited, Nishuvi, Ground Floor, Worli, Mumbai - 400018, India
sucharita.atha@gmail.com
[2]University of Arizona, Tucson, AZ 85721, United States
bolla111@gmail.com

**ABSTRACT**  Foreign Exchange (FOREX) is a decentralised global market for exchanging currencies. The Forex market is enormous, and it operates 24 hours a day. Along with country-specific factors, Forex trading is influenced by cross-country ties and a variety of global events. Recent pandemic scenarios such as COVID19 and local elections can also have a significant impact on market pricing. We tested and compared various predictions with external elements such as news items in this work. Additionally, we compared classical machine learning methods to deep learning algorithms. We also added sentiment features from news headlines using NLP-based word embeddings and compared the performance. Our results indicate that simple regression model like linear, SGD, and Bagged performed better than deep learning models such as LSTM and RNN for single-step forecastings like the next two hours, the next day, and seven days. Surprisingly, news articles failed to improve the predictions indicating domain-based and relevant information only adds value. Among the text vectorization techniques, Word2Vec and SentenceBERT perform better.

## 1   Introduction

Foreign exchange is one of the world's largest markets. Forex trading is available twenty-four hours a day, five days a week. Forex trading occurs over the counter (OTC), which means that it takes place across computer networks. Dailyforex.com [1] reports that daily activity surged to US$ 6.6 trillion in 2019, while the worldwide Forex market

is now worth $2.4 quadrillion. The forex market is around three times the size of the derivatives market and roughly twenty-five times the stock/equity market size. Forecasting foreign currency rates enables brokers and organisations to make informed decisions that avoid risk and increase profits.

Algorithmic trading [1] refers to trading in which, rather than human traders, automated pre-programmed algorithms execute orders. Hedge funds, pension funds, and a variety of other financial institutions employ algorithmic trading. Numerous studies have attempted to forecast the stock market's movement, while relatively few are conducted to forecast the FX market's movement. Global events also have a significant impact on both the stock market and the foreign exchange market. Research has also been done using sentiment analysis of the news headlines and their impact on the stock and forex movement [2].

Forecasting has been studied extensively using both classic machine learning and deep learning models. However, very few studies comprehensively compared models and even fewer using the same dataset. Our work is one of the very few studies that compare conventional models and recent deep learning models. In addition, we compared the effect of external elements such as news headlines on the prediction pattern and examined the improvement that they can make. The outcomes of this paper will guide future researchers and provides insights that enable us to choose the optimal model for Forex prediction.

## 2      Literature Review

Numerous studies have evaluated statistical models such as ARIMA and SARIMA compared with various regression models such as SVR and SVM (wavelet model). The SVR wavelet models outperformed the rest [3]. In a different study, specific hybrid models, ARIMA combined with Fuzzy ARIMA, outperformed simple models [4]. Due to the unpredictable nature of Forex prices, chaos models or a combination of chaos and SVR models also performed well [5]. Later with the advent of deep learning architectures, review papers were published showing the exponential rise in deep learning modeling in the finance domain [6][7]. A comparison was made between ANN models and traditional time series models such as ARIMA, and ANN models performed far better than ARIMA [8]. When comparison was made between CNN and LSTM models for stock predictions, CNN execution time was less, but the results of both the models were comparable. Execution time also depended on the number of variables [9]. Several combination algorithms have also been researched, and they have been shown to outperform the basic models [10][11].

Researchers attempted to incorporate external events into Forex forecasting, as global events affect forex and rates fluctuate dramatically after a significant event. Researches included text data in Forex prediction [12], [13], and observed performance boost with text data. They used classification methodology with TF-IDF and GloVE based vectorization. Researchers have demonstrated that determining the proper sense in powerful words from headlines (Word sense disambiguation) enhances the FOREX movement forecast[14]. One study reveals that market prediction accuracy is higher soon after the news release, implying that news-based trading [15] is the future way.

In the early 2010s, Natural Language Processing (NLP) studies have begun to use neural networks to make better predictions. Word2vec [16] and GloVe [17] sowed the seeds of pre-trained models on a vast corpus and introduced new text vectorization techniques that sparked the transformation of the field of NLP.

In 2017, the Transformer architecture based on attention mechanism achieved significant breakthroughs in several NLP tasks. We can see the better results are obtained when Transformer based modeling is combined with LSTM architecture [18]. Towards the end of 2018 [19], Google researchers developed a new Bidirectional Transformer approach called BERT (Bidirectional Encoder Representations from Transformers), demonstrating significant improvements when applying sentiment analysis in word embeddings. A novel BERT-based Hierarchical Aggregation Model (BHAM) was used to summarize a significant volume of financial news to anticipate forex movement, category-based news outperformed all baselines and all three grouping approaches (time, topic, and category) [20]. Later, researchers introduced FinBERT, a language model based on BERT for financial NLP tasks. FinBERT achieved better performance on two financial sentiment analysis datasets (FiQA sentiment scoring and Financial PhraseBank) [21]. In another study, FinBERT based model derived sentiment from high-frequency text has shown predictive power for Forex price movements [22].

Most previous publications evaluated a single or an ensemble of models and occasionally more than three models. However, there has not been a complete study of forex prediction in the presence of external data such as news headlines. In this work, we have primarily addressed these two issues and focused on improving existing procedures to find the optimum tuning strategies.

## 3 Research Methodology

In this research, we use regression analysis to predict forex prices. The feature "Close_Bid" price is the dependent variable or y, and the other parameters are independent variables. We considered t-1(previous 2 hr), t-12 (previous day), and t-84 (previous week) as independent variables to predict the Close Bid Price.
We performed prediction in two main ways:

- Prediction with only market rates such as Close, Open, High, and Low prices, both univariate and multivariate.
- Prediction with market-rate features (Close, Open, High, and Low prices) and news headlines information. (Fig. 1)

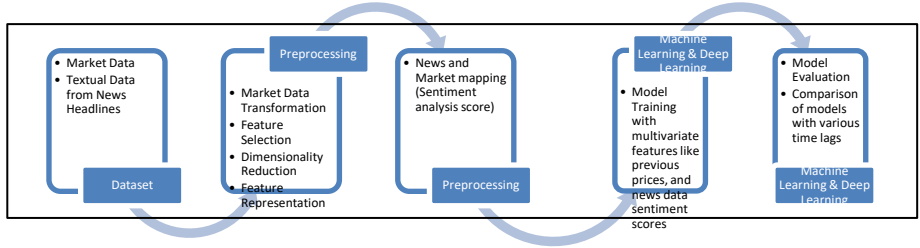

**Fig. 1.** - Architecture for Forex Prediction

**Data Preparation** - The data set includes eight rates for every two hours in a day. If any data is missing, the day's mean value is used to fill in the gaps. Following that, we combine data from the previous two and four hours to create a two-hour timeframe. Additional features are generated as below:

- **Average Price** – The average closing price over the last 5, 30, 90, and 365 days
- **Return** – The return denotes the return on investment. Return = (Price on a particular day – Price on investment day)/ Price on Investment Day. We took this return and derived it for 1, 5, 30, 90, and 365-time intervals.
- **Average of Return** – As the name implies, this is an average of the returns over 5-, 30-, 90-, and 365-day intervals.

We converted the text data to word vectors for the news data. The various word embeddings that we use are – Word2Vec, BERT, and FinBERT.

For Word2Vec[16], we converted the words to vectors, using a feature size of twenty and a context size of ten tokens. In total, we took the first ten features, averaged them over all the tokens. The vectorized features are derived using the Gensim library.

We derived sentence-level embeddings from the Sentence Transformers package, which in turn utilizes BERT-based embeddings ("para-distilroberta-base-v1) as the backbone. For each input sentence, this framework creates embedding. Sentence Transformers changes the pre-trained BERT model by employing Siamese and triplet network topologies [23]. The sum and average of all the dimensions are concatenated with the primary data set as features.

We constructed the final regression model by adding attributes such as textblob polarity score and FinBERT sentiment score. Data was divided into Train, Validation, and Test set. The model was derived using the Train set, and we tuned it using the Validation set. Final results for evaluation were taken based on the Test set.

Below are the various models used for predictions on the top of the prepared dataset:

1. Basic models like - ARIMA, SARIMA, Regression models like Linear Regression, SGD Regressor, XG Boost Regressor, and Support Vector Regressor.
2. Deep learning models like RNN (Recurrent Neural Network), LSTM (Long Short-Term Memory), and GRU (Gated Recurrent Unit)

## 4  Results

### 4.1  Data Sourcing:

The data come from a prior work [12], which can be accessed at the following URL: https://sites.google.com/site/bigdatasetmining/Projects/textmining. The data was primarily derived from two sources: financial markets and news headlines. The market data has EUR/USD market rates for roughly five years, from November 2007 to September 2012. The "market data" comprises features that contain standard market data with the Ask and Bid prices for the currency pairs. The second dataset contains the headlines for the news for each time interval – "News text" – "Time Stamp for News to correspond to two-hour pricing intervals".

### 4.2  Evaluation of models without news data:

SARIMA resulted in a better and comparable score than ARIMA since the data had a seasonal trend (Table 1)

**Table 1.** Timeseries model results

| Time series Model | RMSE | MAE | R2 Score |
|---|---|---|---|
| ARIMA | 0.088 | 0.080 | -2.888 |
| SARIMA | 0.006 | 0.005 | 0.979 |

The performance of regression models is presented in Table 2. Models such as linear regression, SGD, and Bagging regressor have performed well, especially when the lag is lesser. It is interesting to note that deep learning models such as LSTM, GRU, and RNN (Table 2) didn't perform better than machine learning models

**Table 2.** Regression and DL model scores for multiple lags

| Model | RMSE - 2hr lag | MAE - 2hr lag | RMSE - 1day lag | MAE - 1day lag | RMSE - 7day lag | MAE - 7day lag |
|---|---|---|---|---|---|---|
| Linear Regression | **0.0025** | **0.0017** | **0.0036** | **0.0027** | 0.0234 | 0.0187 |
| Bagging Regressor | **0.0025** | **0.0017** | 0.0036 | 0.0028 | 0.0234 | 0.0187 |
| SGD Regressor | 0.0026 | 0.0018 | 0.0088 | 0.0068 | **0.0232** | **0.0187** |
| XGB Regressor | 0.0028 | 0.0020 | 0.0097 | 0.0072 | 0.0247 | 0.0192 |
| Random Forest Regressor | 0.0028 | 0.0020 | 0.0057 | 0.0044 | 0.0291 | 0.0215 |
| Support Vector Regressor | 0.0394 | 0.0335 | 0.0369 | 0.0314 | 0.0453 | 0.0389 |
| LSTM Univariate | 0.0295 | 0.0241 | NA | NA | NA | NA |
| LSTM Multivariate | 0.0120 | 0.0101 | 0.0133 | 0.0228 | 0.0484 | 0.0449 |
| RNN Multivariate | 0.0119 | 0.0097 | 0.0087 | 0.0071 | 0.0316 | 0.0228 |
| GRU Multivariate | 0.0171 | 0.0150 | 0.0301 | 0.0286 | 0.0275 | 0.0253 |

In Linear Regression, we see that the max coefficient is for the previous close price, but when the lag increases the coefficients increase appear for the moving_avg or

avg_price when the lag increases. It varies slightly for 2-h and 7-day lag, but the essential feature always remains the previous closing price (Table 3). While implementing the XGB Regressor, we extracted important features as depicted in Fig. 2. With XGB Regressor as well, the most important feature is the previous Close price, followed by the previous low and high prices.

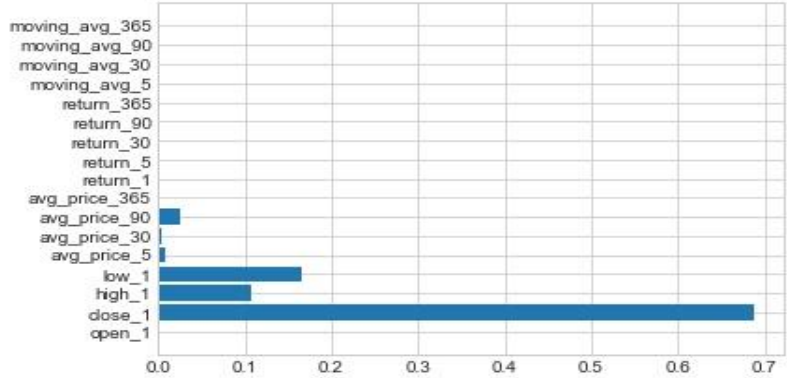

**Fig. 2.** Feature importance using XGB for 1-day lag

**Table 3.** Coefficients for Linear Regression

| Feature | Coefficients - 2hr lag | Coefficients - 1day lag | Coefficients - 7day lag |
|---|---|---|---|
| open_1 | -0.021 | -0.046 | 0.003 |
| close_1 | **0.116** | **0.148** | **0.162** |
| high_1 | -0.001 | -0.011 | -0.127 |
| low_1 | 0.002 | 0.012 | 0.074 |
| avg_price_5 | 0.000 | -0.001 | 0.008 |
| avg_price_30 | -0.002 | -0.020 | -0.027 |
| avg_price_90 | 0.001 | 0.015 | -0.004 |
| avg_price_365 | 0.000 | -0.002 | 0.001 |
| return_1 | -0.001 | -0.002 | -0.001 |
| return_5 | 0.000 | 0.000 | 0.000 |
| return_30 | 0.000 | -0.001 | **-0.001** |
| return_90 | 0.000 | 0.002 | **-0.002** |
| return_365 | 0.000 | 0.000 | **0.001** |
| moving_avg_5 | 0.000 | 0.000 | 0.000 |
| moving_avg_30 | 0.000 | 0.000 | **0.001** |
| moving_avg_90 | 0.000 | -0.001 | **0.003** |
| moving_avg_365 | 0.000 | 0.000 | **-0.001** |

RNN, LSTM, and GRU have performed well among the Deep Learning models, and they, too, have a performance deterioration with an increase in lag, although not as much as the basic regression models. The LSTM Univariate model has not performed as well as multivariate models where time is the variable parameter.

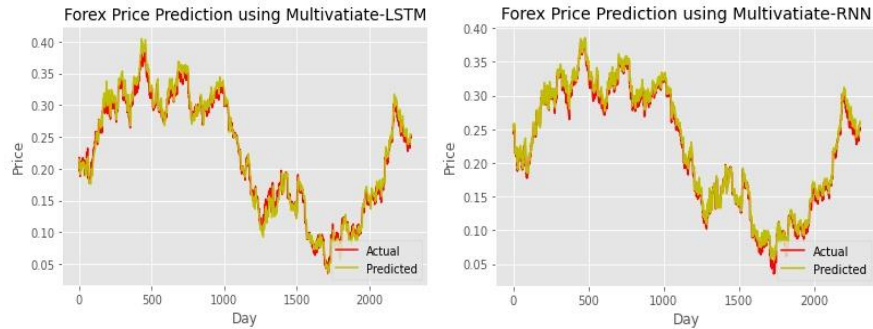

**Fig. 3.** Forex prediction with LSTM and RNN

Linear Regression or SGD Regressor performs as well as or better than LSTM or RNN for all the various lags in predictions. SVR does not produce a satisfactory outcome. When the lag is increased to seven days, we see a degradation in performance, but the basic Regression models still perform better. Another noteworthy finding is that LSTM does not outperform RNN or GRU (Fig. 3). This indicates that memorizing a large amount of prior information may be unnecessary but only a portion of it is needed.

With increasing day lag, we can see that linear regression has a decreasing trend. While RNN scores perform somewhat better with 1-day lag data and decline, they do not decline as much as Linear regression. In a 7-day lag, SGD Regressor performs the best.

### 4.3  Evaluation of models with news data:

Next, we included the news headlines and vectorized them using various word embedding algorithms. Word2Vec, BERT (SentenceBERT), and FinBERT are the tools we utilize. We utilize the various embeddings using two different forms of Machine Learning, like Linear Regression and SVR, two different types of Deep Learning models like LSTM and RNN, and two different lag periods of two hours and one day.

When comparing the outcomes of word embeddings, we observe that both Word2Vec and SentenceBert perform better than FinBERT, implying that features extracted from the text might be more valuable than the sentiment and polarity. Fig 4 depicts a lower sentiment score in some cases followed by lower forex price but is not always consistent. We also observe (Table 4) that models fare better at predicting a 2-hour lag than one day lag with text, and this phenomenon is in line with our observation without using text data. A lower RMSE score denotes a better model.

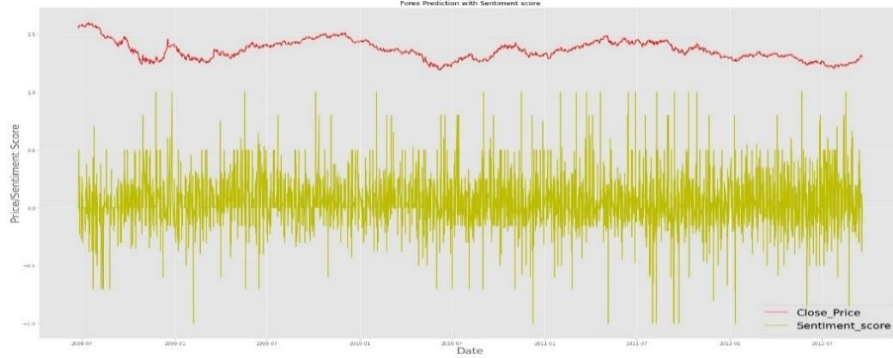

**Fig. 4.** Forex rate (EUR/USD) in red with sentiment score in yellow

| Word Embedding | Model | RMSE - 2hr lag | MAE - 2hr lag | RMSE - 1day lag | MAE - 1day lag |
|---|---|---|---|---|---|
| Word2Vec | Linear Regression | **0.0033** | **0.0022** | **0.0080** | **0.0062** |
|  | Support Vector Regressor | 0.0392 | 0.0332 | 0.0378 | 0.0307 |
|  | LSTM Multivariate | 0.0169 | 0.0110 | 0.0513 | 0.0429 |
|  | RNN Multivariate | 0.0133 | 0.0100 | 0.0462 | 0.0361 |
| Sentence-BERT | Linear Regression | **0.0033** | **0.0022** | **0.0047** | **0.0035** |
|  | Support Vector Regressor | 0.0494 | 0.0439 | 0.0477 | 0.0428 |
|  | LSTM Multivariate | 0.0200 | 0.0253 | 0.0367 | 0.0659 |
|  | RNN Multivariate | 0.0380 | 0.0347 | 0.0305 | 0.0285 |
| FinBERT | Linear Regression | **0.0033** | **0.0021** | **0.0080** | **0.0063** |
|  | Support Vector Regressor | 0.0398 | 0.0336 | 0.0424 | 0.0357 |
|  | LSTM Multivariate | 0.0432 | 0.0529 | 0.0897 | 0.0756 |
|  | RNN Multivariate | 0.0208 | 0.0108 | 0.0655 | 0.0536 |

**Table 4.** RMSE and MAE Score of word embeddings on various models

The RMSE and MAE scores of several models are shown in Table 4, along with their embeddings. In general, we see that linear regression works better with different types of embeddings. We did not evaluate with a further lag of 1 week or more since the results were not as good overall compared to without news data—adding additional latency to the data and seeing how the model works would be a step forward.

According to the study, for various time lags, simple machine learning regression models are preferable. Additionally, the predictive power of Average Returns and moving average closing prices increases as the time lag increases. Furthermore, we observe no effect of external news vectors on prediction performance. One possible explanation is that due to the averaging out the extracting dimensions. Another consideration is the news data's relevance and whether or not all of the data is financial or economic. News about sports or entertainment will not assist in forecasting currency prices.

## 5    Conclusion

Below are some conclusions that we have derived from this work and some future recommendations:

**1.** Overall, we see that the basic regression models such as Linear Regression, Bagging Regressor, and SGD Regressor perform better than the complex deep learning models.

**2.** We observed no improvement in performance when sentiment scores or word embeddings were added to existing data. Few factors may have led to less finance-specific news and fewer news data, which decreased the training set.

**3.** Unlike other studies, ours found no benefit from domain-based embeddings like FinBERT. Further research indicated that the news data was not purely financial, rendering FinBERT embeddings meaningless. The impact of news varies based on the type; this must be evaluated and included as an enhancement.

## References

[1] R. Kissell, *Algorithmic Trading Methods*, 2nd ed. Academic Press, 2020.

[2] N. K. Singh, D. S. Tomar, and A. K. Sangaiah, "Sentiment analysis: a review and comparative analysis over social media," *J. Ambient Intell. Humaniz. Comput.*, vol. 11, no. 1, pp. 97–117, 2020, doi: 10.1007/s12652-018-0862-8.

[3] M. S. Raimundo and J. Okamoto, "SVR-wavelet adaptive model for forecasting financial time series," *2018 Int. Conf. Inf. Comput. Technol. ICICT 2018*, pp. 111–114, 2018, doi: 10.1109/INFOCT.2018.8356851.

[4] F. M. Tseng, G. H. Tzeng, H. C. Yu, and B. J. C. Yuan, "Fuzzy ARIMA model for forecasting the foreign exchange market," *Fuzzy Sets Syst.*, vol. 118, no. 1, pp. 9–19, 2001, doi: 10.1016/S0165-0114(98)00286-3.

[5] S. C. Huang, P. J. Chuang, C. F. Wu, and H. J. Lai, "Chaos-based support vector regressions for exchange rate forecasting," *Expert Syst. Appl.*, vol. 37, no. 12, pp. 8590–8598, 2010, doi: 10.1016/j.eswa.2010.06.001.

[6] Z. Hu, Y. Zhao, and M. Khushi, "A Survey of Forex and Stock Price Prediction Using Deep Learning," *Appl. Syst. Innov.*, vol. 4, no. 1, p. 9, 2021, doi: 10.3390/asi4010009.

[7] O. B. Sezer, M. U. Gudelek, and A. M. Ozbayoglu, "Financial time series forecasting with deep learning: A systematic literature review: 2005–2019," *Appl. Soft Comput. J.*, vol. 90, p. 106181, 2020, doi: 10.1016/j.asoc.2020.106181.

[8] J. Kamruzzaman and R. A. Sarker, "Comparing ANN Based Models with ARIMA for Prediction of Forex Rates," vol. 22, no. 2, pp. 2–11, 2003, [Online]. Available: http://www.asor.org.au/publication/files/jun2003/Joarder.pdf.

[9] J. Sen and S. Mehtab, "Design and Analysis of Robust Deep Learning Models for Stock Price Prediction," 2021, [Online]. Available: http://arxiv.org/abs/2106.09664.

[10] A. Kelotra and P. Pandey, "Stock Market Prediction Using Optimized Deep-ConvLSTM Model," *Big Data*, vol. 8, no. 1, pp. 5–24, 2020, doi:

10.1089/big.2018.0143.

[11] D. K. Nayak and B. K. Bolla, "Efficient Deep Learning Methods for Sarcasm Detection of News Headlines," *Smart Innov. Syst. Technol.*, vol. 269, pp. 371–382, 2022, doi: 10.1007/978-981-16-7996-4_26.

[12] A. Khadjeh Nassirtoussi, S. Aghabozorgi, T. Ying Wah, and D. C. L. Ngo, "Text mining of news-headlines for FOREX market prediction: A Multi-layer Dimension Reduction Algorithm with semantics and sentiment," *Expert Syst. Appl.*, vol. 42, no. 1, pp. 306–324, 2015, doi: 10.1016/j.eswa.2014.08.004.

[13] Y. Liu, J. Trajkovic, H. H. Yeh, and W. Zhang, "Machine Learning for Predicting Stock Market Movement using News Headlines," 2020.

[14] S. Seifollahi and M. Shajari, "Word sense disambiguation application in sentiment analysis of news headlines: an applied approach to FOREX market prediction," *J. Intell. Inf. Syst.*, vol. 52, no. 1, pp. 57–83, 2019, doi: 10.1007/s10844-018-0504-9.

[15] H. N. Semiromi, S. Lessmann, and W. Peters, "North American Journal of Economics and Finance News will tell : Forecasting foreign exchange rates based on news story events in the economy calendar," *North Am. J. Econ. Financ.*, vol. 52, no. February, p. 101181, 2020, doi: 10.1016/j.najef.2020.101181.

[16] T. Mikolov, K. Chen, G. Corrado, and J. Dean, "Efficient estimation of word representations in vector space," *1st Int. Conf. Learn. Represent. ICLR 2013 - Work. Track Proc.*, pp. 1–12, 2013.

[17] J. Pennington, R. Socher, and C. D. Manning, "GloVe: Global Vectors for Word Representation." Accessed: Jan. 15, 2021. [Online]. Available: http://nlp.

[18] F. Eranpurwala, P. Ramane, and B. K. Bolla, "Comparative Study of Marathi Text Classification Using Monolingual and Multilingual Embeddings," *Commun. Comput. Inf. Sci.*, vol. 1534 CCIS, pp. 441–452, 2022, doi: 10.1007/978-3-030-96040-7_35.

[19] J. Devlin, M. W. Chang, K. Lee, and K. Toutanova, "BERT: Pre-training of deep bidirectional transformers for language understanding," *NAACL HLT 2019 - 2019 Conf. North Am. Chapter Assoc. Comput. Linguist. Hum. Lang. Technol. - Proc. Conf.*, vol. 1, no. Mlm, pp. 4171–4186, 2019.

[20] D. Chen, S. Ma, K. Harimoto, R. Bao, Q. Su, and X. Sun, "Group, extract and aggregate: Summarizing a large amount of finance news for forex movement prediction," *arXiv*, 2019, doi: 10.18653/v1/d19-5106.

[21] D. T. Araci, "FinBERT: Financial Sentiment Analysis with Pre-trained Language Models," *arXiv*, 2019.

[22] F. Z. Xing, "High-Frequency News Sentiment and Its Application to Forex Market Prediction."

[23] N. Reimers and I. Gurevych, "Sentence-BERT: Sentence embeddings using siamese BERT-networks," *EMNLP-IJCNLP 2019 - 2019 Conf. Empir. Methods Nat. Lang. Process. 9th Int. Jt. Conf. Nat. Lang. Process. Proc. Conf.*, pp. 3982–3992, 2020, doi: 10.18653/v1/d19-1410.